\documentclass[11pt]{article}
\usepackage[utf8]{inputenc}
\usepackage{amsmath}
\usepackage{amsthm}
\usepackage{amsfonts}
\usepackage{amssymb,mathtools}
\usepackage{latexsym}
\usepackage{bbm}
\usepackage{listings}
\usepackage{graphicx,xcolor}
\usepackage{subfig,caption,booktabs}
\usepackage{multirow}
\usepackage[hidelinks]{hyperref}
\usepackage[toc,page]{appendix}
\usepackage{algorithm}
\usepackage{algorithmic}
\floatname{algorithm}{Algorithm}

\usepackage{siunitx}
\usepackage{authblk}
\usepackage{url}

\newcommand{\rd}{\,\mathrm{d}}
\DeclareMathOperator{\sgn}{sign}
\DeclareMathOperator{\Tr}{Tr}
\DeclareMathOperator{\Hessian}{Hess}

\newcommand{\NN}{\mathfrak{N}}

\makeatletter
\begin{document}

\title{Solving high-dimensional eigenvalue problems using deep neural networks: A diffusion Monte Carlo like approach}

\author[1]{Jiequn Han}
\author[2,3]{Jianfeng Lu}
\author[2]{Mo Zhou} 
\affil[1]{Department of Mathematics, Princeton University}
\affil[2]{Department of Mathematics, Duke University}
\affil[3]{Department of Physics, and Department of Chemistry, Duke University}

\maketitle
\begin{abstract}
We propose a new method to solve eigenvalue problems for linear and semilinear second order differential operators in high dimensions based on deep neural networks. The eigenvalue problem is reformulated as a fixed point problem of the semigroup flow induced by the operator, whose solution can be represented by Feynman-Kac formula in terms of forward-backward stochastic differential equations. The method shares a similar spirit with diffusion Monte Carlo but augments a direct approximation to the eigenfunction through neural-network ansatz. The criterion of fixed point provides a natural loss function to search for parameters via optimization. Our approach is able to provide accurate eigenvalue and eigenfunction approximations in several numerical examples, including Fokker-Planck operator and the linear and nonlinear Schr\"odinger operators in high dimensions.
\end{abstract}

\section{Introduction}
Many fundamental problems in scientific computing can be reduced to the computation of eigenvalues and eigenfunctions of an operator. One primary example is the electronic structure calculations, namely, computing the leading eigenvalue and eigenfunction of the Schr\"odinger operator. 
If the dimension of the state variable is low, one can use classical approaches, such as the finite difference method or spectral method, to discretize the operator and to solve the eigenvalue problem. However, these conventional, deterministic approaches suffer from the so-called curse of dimensionality, when the underlying dimension becomes high, since the number of degrees of freedom grows exponentially as the dimension increases. 

For high-dimensional problems, commonly arising from quantum mechanics, statistical mechanics, and finance applications, stochastic methods become more attractive and in many situations the only viable option. 
In the context of quantum mechanics, two widely used approaches for high-dimensional eigenvalue problems are the variational Monte Carlo (VMC) and diffusion Monte Carlo (DMC) methods~\cite{mcmillan1965VMC,ceperley1977VMC,blankenbecler1981AFQMC,zhang1997AFQMC,foulkes2001QMC,needs2009continuum}. These two approaches deal with the high dimensionality via different strategies. 
VMC relies on leveraging chemical knowledge to propose an ansatz of the eigenfunction (wavefunction in the context of quantum mechanics) with parameters to be optimized under the variational formulation of the eigenvalue problem. The Monte Carlo approach is used to approximate the gradient of the energy with respect to the parameters at each optimization iteration step. On the other hand, DMC represents the density of the eigenfunction with a collection of particles that follow the imaginary time evolution given by the Schr\"odinger operator, via a  Feynman-Kac representation of the semigroup. It can be understood as a generalization of the classical power method  from finite-dimensional matrices to infinite-dimensional operators. In electronic structure calculations, DMC usually can give more accurate eigenvalues compared with VMC, which relies on the quality of the proposed ansatz, while the particle representation of DMC often falls short of providing other information of the eigenfunction, such as its derivatives, unlike VMC.

As discussed above, one key to solving high-dimensional eigenvalue problems is the choice of function approximation to the targeted eigenfunction, ranging from the grid-based basis, spectral basis, to nonlinear parametrizations used in VMC, and to particle representations in DMC. Given the recent compelling success of neural networks in representing high-dimensional functions with remarkable accuracy and efficiency in various computational disciplines, it is fairly attempting to introduce neural networks to solve high-dimensional eigenvalue problems.
This idea has been recently investigated under the variational formulation by \cite{E2018deep,Pfau2018spectral,Han2019solving,Pfau2019abinitio,Hermann2019deep,Choo2019fermionic}.
Particularly \cite{Han2019solving,Pfau2019abinitio,Hermann2019deep,Choo2019fermionic} has shown the exciting potential of solving the many-electron Schr{\"o}dinger equation with neural networks within the framework of VMC. 
On the other hand, how to apply neural networks in the formalism of DMC has not been explored in the literature, which leaves a natural open direction to investigate.

In this paper, we propose a new algorithm to solve high-dimensional eigenvalue problem for the second-order differential operators, in a similar spirit of DMC while based on the neural network parametrization of the eigenfunction.
The eigenvalue problem is reformulated as a parabolic equation, whose solution can be represented by (nonlinear) Feynman-Kac formula in terms of forward-backward stochastic differential equations. Then we leverage the recently proposed deep BSDE method \cite{E2017deep, Han2018solving} to seek optimal eigenpairs. 
Specifically, two deep neural networks are constructed to represent the eigenfunction and its scaled gradient. Then the neural network is propagated according to the semigroup generated by the operator. The loss function is defined as the difference between the neural networks before and after the propagation.
Compared to conventional DMC, the proposed algorithm provides a direct approximation to the target eigenfunction, which overcomes the shortcoming in providing the gradient information.
Moreover, since the BSDE formulation is valid for nonlinear operators, our approach can be extended to high-dimensional nonlinear eigenvalue problems, as validated in our numerical examples.

The rest of this paper is organized as follows. In Section 2, we introduce the algorithm to solve the eigenvalue problem. In Section 3, numerical examples are presented. We conclude in Section 4 with an outlook for future work.

\section{Numerical methods}

\subsection{The method for a linear operator}
We consider the eigenvalue problem 
\begin{equation}
\label{eqn:eigen_prob}
\mathcal{L}\psi = \lambda \psi,
\end{equation}
on $\Omega = [0,2\pi]^d$ with periodic boundary condition where $\mathcal{L}$ is a linear operator of the form
\begin{equation}
\label{eqn:operator}
\mathcal{L} \psi(x) = -\frac{1}{2} \Tr \left( \sigma \sigma^{\top} \Hessian(\psi)(x) \right) - b(x) \cdot \nabla \psi(x) + f(x) \psi(x).
\end{equation}
$\sigma$ is a $d \times d$ {constant invertible} matrix such that $\sigma \sigma^{\top}$ is positive definite, $\nabla \psi$ denotes the gradient of $\psi$, $b(x)$ is a $d$-dimensional vector field and $\Hessian(\psi)$ denotes the Hessian matrix of $\psi$.  

To solve this eigenvalue problem, we augment a time variable and consider the following backward parabolic partial differential equation (PDE) in the time interval $[0, T]$:
\begin{equation}\label{eqn:BackPDE}
\left\{ \begin{aligned}
\partial_t u(t,x) - \mathcal{L} u(t,x) + \lambda u(t,x) & = 0 ~~~ & \text{in} ~ [0,T] \times \Omega,\\
u(T,x) & = \Psi(x) ~~~ & \text{on} ~ \Omega.
\end{aligned} \right.
\end{equation}
This is essentially a continuous time analog of the power iteration for
matrix eigenvalue problem. Let us denote the solution of
\eqref{eqn:BackPDE} as
$u(T-t,\cdot) = \mathcal{P}^{\lambda}_{t} \Psi$ (note that the backward propagator
$\{\mathcal{P}^{\lambda}_t\}_{t\leq T}$ forms a semigroup, i.e., $\mathcal{P}^{\lambda}_{t_1} \circ \mathcal{P}^{\lambda}_{t_2} = \mathcal{P}^{\lambda}_{t_1 + t_2}$). 
According to the spectral theory of the elliptic operator, if $\Psi$
is a stationary solution of~\eqref{eqn:BackPDE}, i.e.,
$\mathcal{P}_T^\lambda\Psi = \Psi$, then $(\lambda, \Psi)$ must be an
eigenpair of $\mathcal{L}$.  Therefore, we can minimize the ``loss
function'' $\|\mathcal{P}_T^\lambda\Psi- \Psi\|^2$ with respect to
$(\lambda, \Psi)$ to solve the eigenvalue problem. While this is a non-convex optimization problem, we expect local convergence to a valid eigenpair with appropriate initialization. 
{Note that, unlike power iteration in which $\lambda$ is determined by the eigenvector, in our algorithm $\lambda$ is treated as a variational parameter and optimized jointly with the eigenfunction, as in the DMC method.
}

The reformulation above turns the eigenvalue problem into solving a parabolic PDE in high dimensions. For the latter, we can leverage the recently developed deep BSDE method~\cite{E2017deep,Han2018solving,Han2018convergence} (which is why the parabolic PDE \eqref{eqn:BackPDE} is written backward in time). Let $X_t$ solve the stochastic differential
equation (SDE) 
\begin{equation}\label{SDE1}
\rd X_t = \sigma \rd W_t,
\end{equation}
or in the integral form
\begin{equation}\label{eqn:Ito_Xt}
X_t = X_0 + \int_{0}^{t} \sigma \rd W_s,
\end{equation}
where $W_t$ is a $d$-dimensional Brownian motion, and $X_0$ is sampled from some initial distribution $\nu$. Then according to the It\^o's
formula, the solution to \eqref{eqn:BackPDE}, $u(t, x)$ satisfies
\begin{equation}\label{eqn:Ito_uXt}
u(t,X_t) = u(0,X_0) + \int_{0}^{t} (f(X_s)u(s,X_s) - \lambda u(s,X_s) - b(X_s) \nabla u(s,X_s) ) \rd s + \int_{0}^{t} \sigma^{\top} \nabla u(s, X_s) \rd W_s.
\end{equation}
Note that simulating the two SDEs \eqref{eqn:Ito_Xt} and
\eqref{eqn:Ito_uXt} is relatively simple even in high dimensions,
while directly solving the PDE \eqref{eqn:BackPDE} is intractable. We remark that it is possible to add a drift term $b(X_t) \rd t$ to the SDE \eqref{SDE1} and modify \eqref{eqn:Ito_uXt} accordingly (see the discussion below). 

Of course, a priori in
\eqref{eqn:Ito_uXt} for both $u(s, \cdot)$ and $\nabla u(s, \cdot)$ are
unknown, while we know that if we set
$u(s, \cdot) = \Psi(\cdot)$, the eigenfunction we look for, and $\nabla u(s, \cdot) = \nabla \Psi(\cdot)$, the solution $u(t, \cdot)$ remains $\Psi(\cdot)$ for all $t$. The idea of our method is then to use two neural networks, $\NN_{\Psi}$ and
$\NN_{\sigma^{\top} \nabla \Psi}$ as ansatz for the eigenfunction
$\Psi$ and its scaled gradient $\sigma^{\top} \nabla \Psi$,
respectively. Assigning $u(0, X_0)=\NN_{\Psi}(X_0)$ and
$ \sigma^{\top} \nabla u(s, X_s) =\NN_{\sigma^{\top} \nabla
  \Psi}(X_s)$
in \eqref{eqn:Ito_uXt}, the discrepancy for the propagated solution,
i.e.,
\begin{equation}
\label{eqn:consistency}
\mathbb{E}_{X_0 \sim \nu} \Bigl[ \eta_1\bigl\lvert \NN_{\Psi}(X_T) - u(T, X_T) \bigr\rvert^2 + \eta_2 \bigl\lvert \NN_{\sigma^{\top}\nabla \Psi}(X_T) - \sigma^{\top}\nabla \NN_{\Psi}(X_T) \bigr\rvert^2  \Bigr]
\end{equation}
then indicates the accuracy of the approximation. Here, we use $|\cdot|$ to denote the absolute value of a number or the Euclidean norm of a vector according to the context. Note that the second
term above penalizes the discrepancy between the approximation of
$\Psi$ and its gradient, where $\eta_1,\eta_2$ are two weight
hyperparameters. Therefore, using the above discrepancy as a loss function
to optimize the triple
$(\lambda, \NN_{\Psi}, \NN_{\sigma^{\top}\nabla \Psi})$ gives us a
scheme to solve the eigenvalue problem. The above procedure can be
directly extended to semilinear case where $f$ depends on $\Psi$ and
$\nabla \Psi$, as we will discuss in Section \ref{sec:semilinear}.

To employ the above framework in practice, we numerically discretize
the SDEs \eqref{eqn:Ito_Xt} and \eqref{eqn:Ito_uXt} using
Euler--Maruyama method with a given partition of interval $[0,T]$ :
$0 = t_0 < t_1 < \cdots < t_N = T$:
\begin{align}\label{eqn:discrete_X}
& \mathcal{X}_0 = X_0, &&
\mathcal{X}_{t_{n+1}} = \mathcal{X}_{t_n} + \sigma \Delta W_n,
\intertext{and}
\label{eqn:discrete_u}
& \mathcal{U}_0 = \NN_{\Psi}(\mathcal{X}_0), &&
\mathcal{U}_{t_{n+1}} = \mathcal{U}_{t_n} + \left(f(\mathcal{X}_{t_n})\mathcal{U}_{t_n} - \lambda \mathcal{U}_{t_n} - (b \sigma^{-1 \top} \NN_{\sigma^{\top}\nabla \Psi})(\mathcal{X}_{t_n})\right) \Delta t_n + \NN_{\sigma^{\top}\nabla \Psi}(\mathcal{X}_{t_n}) \Delta W_n,
\end{align}
for $n = 0, 1, \ldots, N-1$.
Here $\Delta t_n = t_{n+1} - t_{n}$, $\Delta W_n = W_{t_{n+1}} - W_{t_n}$, and we use $X_t$ and $\mathcal{X}_t/\mathcal{U}_t$ to represent the continuous and discretized stochastic process, respectively. The noise terms $\Delta W_n$ have the same realization in \eqref{eqn:discrete_X} and \eqref{eqn:discrete_u}, as in the forward-backward SDEs \eqref{eqn:Ito_Xt} and \eqref{eqn:Ito_uXt}. 

The loss function \eqref{eqn:consistency} then corresponds to the discrete counterpart:  
\begin{equation}
\label{eqn:discrete_loss}
\mathbb{E}_{X_0 \sim \nu}\Bigl[\eta_1\bigl\lvert\NN_{\Psi}(\mathcal{X}_T) - \mathcal{U}_T\bigr\rvert^2 + \eta_2 \bigl\lvert \NN_{\sigma^{\top} \nabla \Psi}(\mathcal{X}_T) - \sigma^{\top} \nabla \NN_{\Psi}(\mathcal{X}_T) \bigr\rvert^2\Bigr],
\end{equation}
where $\nabla \NN_{\Psi}$ is the gradient of neural network $\NN_{\Psi}$ with respect to its input. 
In practice, the expectation in \eqref{eqn:discrete_loss} is further approximated by Monte Carlo sampling, which is similar to the empirical loss often used in the supervised learning context. For a given batch size $K$, we sample $K$ points $\{\mathcal{X}_0^k\}_{k=1}^K$ of the initial state from the distribution $\nu$ at each training step and estimate the gradient of the loss with respect to the trainable parameters using the empirical Monte Carlo average of \eqref{eqn:discrete_loss}:
\begin{equation}
\label{eqn:MC_loss}
\frac{1}{K} \sum_{k=1}^K \,\Bigl[\eta_1\bigl\lvert \NN_{\Psi}(\mathcal{X}_T^k) - \mathcal{U}_T^k\bigr\rvert^2 + \eta_2 \bigl\lvert \NN_{\sigma^{\top} \nabla \Psi}(\mathcal{X}_T^k) - \sigma^{\top} \nabla \NN_{\Psi}(\mathcal{X}_T^k) \bigr\rvert^2\Bigr].
\end{equation}
We remark that the definition of the dynamic \eqref{eqn:Ito_Xt} is not unique and implicitly affects the detailed computation of the loss function \eqref{eqn:discrete_loss} and \eqref{eqn:MC_loss}. Specifically, in  \eqref{eqn:Ito_Xt}, the diffusion term $\sigma$ is determined by the operator \eqref{eqn:operator} while the choice of initial distribution and the drift term has some flexibility.  If the drift in \eqref{eqn:Ito_Xt} changes, one can change the associated drift in \eqref{eqn:Ito_uXt} according to It\^o's
formula and define the loss again as \eqref{eqn:discrete_loss} and \eqref{eqn:MC_loss}. In this work, we choose the form of \eqref{eqn:Ito_Xt} without the drift and $\nu$ being the uniform distribution on $\Omega$ to ensure that the whole region is reasonably sampled for the optimization of eigenfunction. Some importance sampling can also be used if some prior knowledge of the eigenfunction is available, which we will not go into further details in this work.

At a high level, our algorithm is in a similar vein as the power method for solving the eigenvalue problem in linear algebra. Both algorithms seek for a solution that is stationary under the propagation. However, one distinction is that our algorithm  {may also be used for general eigenvalues depending on the initialization of $\lambda$ and $\Psi$. Using matrix notation, this is similar to using the objective function $\lVert(A - \lambda) v\rVert^2$ to find non-dominant eigenpair $\lambda$ and $v$ for the matrix $A$. The actual performance of solving for non-dominant eigenvalue depends on the initialization and spectral gap between eigenvalues, as will be illustrated by numerical results in Section 3.}
On the other hand, we find in the numerical experiments that if $\lambda$ is initialized small enough, it will always converge to the first eigenvalue.

\begin{figure}\centering 
\includegraphics[width=0.98\textwidth]{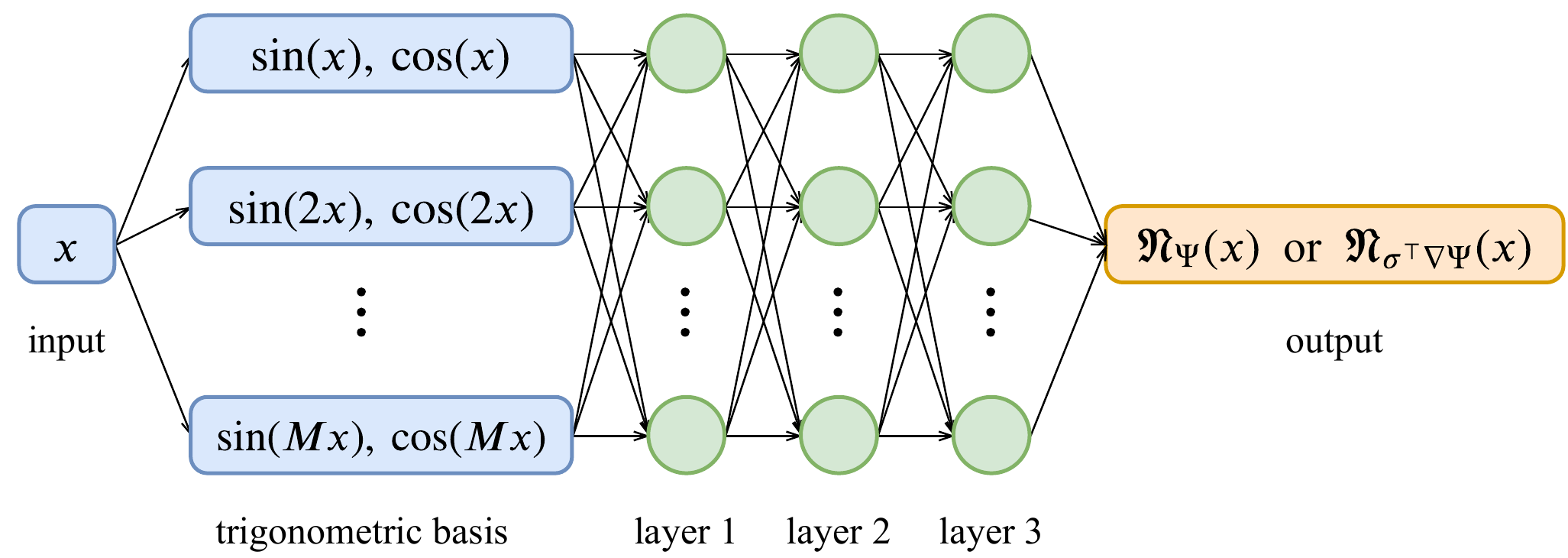}
\caption{Illustration of the neural network with $3$ layers and several nodes in each layer. For $x=(x_1,\cdots,x_d)$, $\sin(kx) \coloneqq (\sin(kx_1), \sin(kx_2), \cdots, \sin(kx_d))$ is a $d$-dimensional vector, and $\cos(kx)$ is defined similarly, $k=1,2,\cdots, M$.  The output for $\NN_{\Psi}$ is a scalar and the output for $\NN_{\sigma^{\top} \nabla \Psi}$ is a $d$-dimensional vector.}
\label{fig:NN}
\end{figure}

In practice, we use fully-connected feed-forward neural networks for the approximation of $\Psi$ and $\sigma^{\top}\nabla \Psi$, respectively. To ensure periodicity of the neural network outputs, the input vector $x=(x_1,\dots,x_d)$ is first mapped into a fixed trigonometric basis $\{\sin(jx_i), \cos(jx_i) \}_{i=1,j=1}^{~d,~~M}$ of order $M$. 
Then the vector consisting of all basis components are fed into fully-connected neural networks with some hidden layers, each with several nodes. See Figure~\ref{fig:NN} for illustration of the involved network structure.  We use ReLU as the activation function and optimize the parameters with the Adam optimizer~\cite{Kingma2015adam}. 

\subsection{Normalization}
The above loss has one caveat though, as the trivial solution 
$(\NN_{ \Psi} =0, \NN_{ \sigma^{\top} \nabla \Psi}=0)$ is a global minimizer. Therefore, normalization is required to exclude such a trivial case. We seek for eigenfunctions $\Psi$ such that $\int_{\Omega} \Psi^2 = |\Omega|$, i.e., $\frac{1}{|\Omega|} \|\Psi \|^2_{L^2} =1$.~\footnote{The reason we set $\frac{1}{|\Omega|} \|\Psi \|^2_{L^2} =1$ instead of $\| \Psi\|_{L^2}^2 = 1$ is because we want to consider the problem in high dimensions in the domain $\Omega = [0,2\pi]^d$. Consider the trivial case when $\mathcal{L} = -\Delta$, whose smallest eigenvalue is $\lambda=0$ and any constant function is a corresponding eigenfunction. If $\| \Psi\|_{L^2}^2 = 1$, the constant function becomes $\Psi = (\frac{1}{2\pi})^{d/2}$, which vanishes as $d \to \infty$; instead the normalization $\lVert \Psi \rVert_{L^2}^2 = \lvert \Omega \rvert$ keeps the pointwise-value of $\Psi$ as order $1$, which benefits the training process.}
To proceed, we define the normalization constant
\begin{equation}
\label{eqn:normalization_const}
Z_\Psi = \sgn \Bigl(\int_\Omega \NN_{\Psi}(x)\rd x\Bigr) \biggl(\frac{1}{|\Omega|} \int_\Omega  \NN_{\Psi}(x)^2 \rd x \biggr)^{1/2}. 
\end{equation}
Thus
dividing $\NN_{\Psi}$ by $Z_\Psi$, we enforce that the parametrized function ensures the normalization condition. Note that the first factor on the right hand side of
\eqref{eqn:normalization_const} is introduced to fix the global sign
ambiguity of the eigenfunction.

In computation, given the parameters of $\NN_{\Psi}$, we do not have direct access to $Z_\Psi$. Instead, at the $\ell$-th step of training, we use our batch of $K$ data samples to approximate $Z_\Psi$ via
\begin{equation} 
\label{eqn:Z_batch}
\hat{Z}^{\ell}_\Psi = \sgn \Bigl(\sum_{k=1}^K \NN_{\Psi}( \mathcal{X}^{k, \ell}_{0})\Bigr) \biggl( \frac1K\sum_{k=1}^{K}  \NN_{\Psi}(\mathcal{X}^{k, \ell}_{0})^2 \biggr)^{1/2}, 
\end{equation}
where the superscripts in $\mathcal{X}^{k, \ell}_0$ serve as the index of batch ($k$) and index of the training step ($\ell$). The above is a Monte Carlo estimation of \eqref{eqn:normalization_const} if $\mathcal{X}_0$ is sampled from the uniform distribution (which we assume in this work). %
Due to the normalization procedure, $\hat{Z}_{\Psi}^\ell$ will enter into the loss function, and thus the stochastic gradient based on the empirical average over the batch becomes biased (since $\mathbb{E} (A / B) \neq \mathbb{E} A / \mathbb{E} B$ in general).
To reduce the bias and make the training more stable, we introduce an exponential moving average scheme to the normalization constant in order to reduce the dependence of the loss on the estimated normalization constant of the current batch.
In our implementation, we use 
\begin{equation} \label{eqn:moving_average}
Z^{\ell}_{\Psi} = \gamma_{\ell} Z^{\ell-1}_{\Psi} + (1 - \gamma_{\ell}) \hat{Z}^{\ell}_{\Psi}.
\end{equation}
Here $\gamma_{\ell} \in (0,1)$ is the moving average coefficient  for decay. It is observed that small $\gamma_{\ell}$ at the beginning makes training efficient, and later on its value is increased such that the gradient is less biased.

Given the introduced normalization factor, the neural network approximation $\mathcal{U}_0$ in the updating scheme \eqref{eqn:discrete_u} is replaced by (we suppress the training step index in $\mathcal{X}$)
\begin{equation}
\label{eqn:normalization1}
\mathcal{U}_0^k = \frac{\NN_{\Psi}(\mathcal{X}^k_{0})}{Z_\Psi^{\ell}},
\end{equation}
and we would hope to reduce the discrepancy between the solution of \eqref{eqn:discrete_u} at time $T$ and the normalized neural network approximation $\NN_{\Psi}(\mathcal{X}^k_T) / Z^{\ell}_\Psi$ through training. The associated batch approximation of loss function used for the computation of stochastic gradient is as follows 
\begin{equation}\label{eqn:loss}
\frac{1}{K}\sum_{k=1}^K \left(  \eta_1\bigg\vert \frac{\NN_{\Psi}(\mathcal{X}^k_T)}{Z^\ell_\Psi} - \mathcal{U}_T^k \bigg\vert^2 + \eta_2 \bigg\vert \NN_{\sigma^{\top} \nabla \Psi}(\mathcal{X}^k_T) -  \frac{\sigma^{\top}\nabla \NN_{\Psi}(\mathcal{X}^k_T)}{Z^\ell_\Psi} \bigg\vert^2 \right)
+ \eta_3(Z_0 - Z^\ell_\Psi)^+.
\end{equation}
In the last term above, $Z_0$ is a hyperparameter and $\eta_3$ is the associated weight. This term is introduce to prevent $Z^\ell_\Psi$ being too small; otherwise the normalization would become unstable. In each training step, we calculate the gradient of \eqref{eqn:loss} with respect to all the parameters to be
optimized, including the eigenvalue $\lambda$ and parameters in the neural
network ansatz $\NN_{\Psi}$ and $\NN_{\sigma^{\top} \nabla \Psi}$.
Note that in \eqref{eqn:loss} we do not normalize $\NN_{\sigma^{\top} \nabla \Psi}$ since its scale has been determined implicitly.
When $K$ is reasonably large and if we neglect the discretization error of simulating the SDEs, the empirical sum in \eqref{eqn:loss} can be interpreted as a Monte Carlo approximation to the loss (ignoring the sign ambiguity)
\begin{equation}\label{eqn:empirical loss}
\mathbb{E}_{X_0 \sim \nu}\left[\eta_1\bigg\vert \frac{\NN_{\Psi}(X_T) }{\|\NN_{\Psi}\|_2 / | \Omega|^{\frac12}} - u(T,X_T) \bigg\vert^2 + \eta_2 \bigg\vert \NN_{\sigma^{\top} \nabla \Psi}(X_T)  - \frac{\sigma^{\top} \nabla \NN_{\Psi}(X_T)}{\|\NN_{\Psi}\|_2 / | \Omega|^{\frac12}} \bigg\vert^2\right],
\end{equation}
where $u(T,X_T)$ is defined as \eqref{eqn:Ito_uXt} except that $u(0,X_0)=| \Omega|^{\frac12}\NN_{\Psi}(X_0)/\|\NN_{\Psi}\|_2 $. 
We remark that the normalization procedure introduced here shares a similar spirit with Batch Normalization \cite{Ioffe2015batch}, which is widely used in the training of neural networks.

We summarize our algorithm as pseudocode in Algorithm~\ref{alg}. 

\begin{algorithm}%
\caption{Neural network based eigensolver}
\begin{algorithmic}
\REQUIRE operator $\mathcal{L}$, terminal time $T$, number of time intervals $N$, loss weights $\eta_1, \eta_2, \eta_3$, $Z_0$, neural network structures, number of iterations, learning rate, batch size $K$, moving average coefficient $\gamma_{\ell}$ in \eqref{eqn:moving_average}
\ENSURE eigenvalue $\lambda$, eigenfunction $\NN_{\Psi}$ and rescaled gradient $\NN_{\sigma^{\top} \nabla \Psi}$
\STATE{initialization: eigenvalue $\lambda$, $\NN_{\Psi}$, $\NN_{\sigma^{\top} \nabla \Psi}$ and normalization factor $Z_{\Psi}^0$}
\FOR{$\ell= 1$ \TO the number of iterations} 
\STATE {sample $K$ points of $\mathcal{X}_0$ and sample $K$ Wiener processes $W_t$} 
\STATE {compute $\mathcal{X}_{t_n}$ via \eqref{eqn:discrete_X}} 
\STATE {compute the normalization factor $Z^\ell_{\Psi}$ via \eqref{eqn:Z_batch} and \eqref{eqn:moving_average}} 
\STATE {normalize and propagate via \eqref{eqn:normalization1} and \eqref{eqn:discrete_u}} 
\STATE {compute the gradient of loss \eqref{eqn:loss} with respect to the trainable parameters} 
\STATE {update the trainable parameters by the Adam method} 
\ENDFOR
\end{algorithmic}
\label{alg}
\end{algorithm}

\subsection{ The method for a semilinear operator}
\label{sec:semilinear}

Our algorithm can be generalized to solve eigenvalue problems for semilinear operator
\begin{equation}
\mathcal{L} \psi(x) = -\frac{1}{2} \Tr \left( \sigma \sigma^{\top} \Hessian(\psi)(x) \right) - b(x) \cdot \nabla \psi(x) + f(x, \psi(x), \sigma^{\top}\nabla \psi(x)).
\end{equation}
The method for semilinear problems is almost the same to previous sections, except for a few modifications. The SDE for $X_t$ is the same as \eqref{eqn:Ito_Xt}
while equation \eqref{eqn:Ito_uXt} that the solution of the PDE \eqref{eqn:BackPDE} satisfies becomes
\begin{equation}
\begin{aligned}
u(t,X_t) = &u(0,X_0) + \int_{0}^{t} \left( f(X_s, u(s,X_s), \sigma^{\top} \nabla u(s,X_s)) - b(X_s) \cdot \nabla u(s,X_s) - \lambda u(s,X_s) \right) \rd s\\
&+ \int_{0}^{t} \sigma^{\top} \nabla u(s, X_s) \rd W_s.
\end{aligned}
\label{eqn:Ito_uXt2}
\end{equation}
The discretization of $X_t$, equation \eqref{eqn:discrete_X}, remains unchanged while equation \eqref{eqn:discrete_u} needs modification according to \eqref{eqn:Ito_uXt2}:
\begin{equation}\label{eqn:discrete_u2}
\begin{aligned} 
\widetilde{\mathcal{U}}_{t_{n+1}} &= \mathcal{U}_{t_n} + \left(f(\mathcal{X}_{t_n}, \mathcal{U}_{t_n}, \NN_{\sigma^{\top}\nabla \Psi}) - \lambda \mathcal{U}_{t_n} - (b \sigma^{-\top} \NN_{\sigma^{\top}\nabla \Psi})(\mathcal{X}_{t_n})\right) \Delta t_n + \NN_{\sigma^{\top}\nabla \Psi}(\mathcal{X}_{t_n}) \Delta W_n,\\
\mathcal{U}_{t_{n+1}} &= \text{Clip}(\widetilde{\mathcal{U}}_{t_{n+1}}, P, Q),
\end{aligned}
\end{equation}
where (for $P<Q$) Clip is a clipping function given by 
\begin{equation} \label{eqn:clip}
\text{Clip}(u,P,Q) = 
\begin{cases} 
P, &\text{if } u < P;\\
u, &\text{if } P \le u \le Q;\\
Q, &\text{if } u > Q.
\end{cases}
\end{equation}
Here we introduce the clipping function ($-P=Q>0$) to prevent numerical instability caused by the nonlinearity of $f$ in \eqref{eqn:Ito_uXt2} 
{(for instance, the cubic term in the nonlinear Schr\"odinger equation \eqref{eq:nlse} below)}, especially at the early stage of training. It checks $\mathcal{U}_{t_{n}}$ and replaces those whose absolute values are larger than $Q$ with $\sgn(\mathcal{U}_{t_{n}}) Q$, where $Q>0$ is an upper bound of the absolute value of the true normalized eigenfunction. 
Given the modified forward dynamics \eqref{eqn:discrete_u2}, the loss function for the semilinear operators are defined the same as \eqref{eqn:loss}, and the training algorithm is the same too. 

\section{Numerical results}
In this section, we report the performance of the proposed eigensolver in three examples: the Fokker-Planck equation, the linear Schr\"odinger equation, and the nonlinear  Schr\"odinger equation. The domain $\Omega$ is always $[0, 2\pi]^d$ with periodic boundary condition. In each example we consider the dimension $d=5$ and $d=10$. {We also solve the second eigenpair of the linear Schr\"odinger equation with $d=10$ to illustrate that our algorithm is able to get non-dominant eigenpairs.} All The hyperparameters are given in~\ref{hyperparameter}.
We examine the errors of the prescribed eigenvalue, the associated eigenfunction, and the gradient of the eigenfunction.
The errors for eigenfunctions and gradients of eigenfunctions are computed in the $L^2$ {and $L^{\infty}$} sense, approximated through a set of validation points. Given a set of  validation points $\{\mathcal{X}^k\}_{k=1}^K$, we use the quantity
\begin{equation}
\text{err}_{\Psi} \coloneqq \left[\frac1K \sum_{k=1}^K \Biggl( \frac{\NN_{\Psi}(\mathcal{X}^k)}{Z^\ell_{\Psi}} - \frac{\Psi(\mathcal{X}^k)}{\bigl( \frac{1}{K} \sum_{m=1}^K \Psi (\mathcal{X}^m)^2 \bigr)^{\frac12}} \Biggr)^2 \right]^{\frac12}
\end{equation}
{and
\begin{equation}
\text{err}_{\Psi}^{\infty} \coloneqq \max_{k} \left|\frac{\NN_{\Psi}(\mathcal{X}^k)}{Z^\ell_{\Psi}} - \frac{\Psi(\mathcal{X}^k)}{\bigl( \frac{1}{K} \sum_{m=1}^K \Psi (\mathcal{X}^m)^2 \bigr)^{\frac12}}\right|
\end{equation}
}
to measure the error for eigenfunction where $Z^\ell_{\Psi}$ is computed via equation \eqref{eqn:moving_average}, with a known reference eigenfunction $\Psi$. We use
\begin{equation}
\text{err}_{\sigma^\top \nabla \Psi} \coloneqq \left[ \frac{1}{Kd} \sum_{k=1}^K \left| \frac{\NN_{\sigma^\top \nabla \Psi} (\mathcal{X}^k)}{\bigl(\frac{1}{Kd}\sum_{m=1}^K | \NN_{\sigma^\top \nabla \Psi} (\mathcal{X}^m) |^2 \bigr)^{\frac12}} - \frac{\sigma^\top \nabla \Psi (\mathcal{X}^k)}{\bigl(\frac{1}{Kd}\sum_{m=1}^K | \sigma^\top\nabla \Psi (\mathcal{X}^m) |^2 \bigr)^{\frac12}} \right|^2 \right]^{\frac12}
\end{equation}
to quantify the error for the gradient approximation. We record and plot the error every $100$ steps in the training process, with a smoothed moving average of window size $10$. The final error reported is based on the average of the last $1000$ steps.
Besides the errors above, we also visualize and compare the density of the true eigenfunction and its neural network approximation (since it is hard to visualize the high-dimensional eigenfunction directly). The density of a function $\Psi$ is defined as the probability density function of $\Psi(X)$ where $X$ is a uniformly distributed random variable on $\Omega$. In practice, the density is approximated by Monte Carlo sampling. As shown below, in all three examples, we find that the eigenpairs (with gradients) are solved accurately and the associated densities match well.

\subsection{Fokker-Planck equation}
\label{secFP}
\begin{figure*}[t!]
    \centering
    \includegraphics[width=0.98\textwidth]{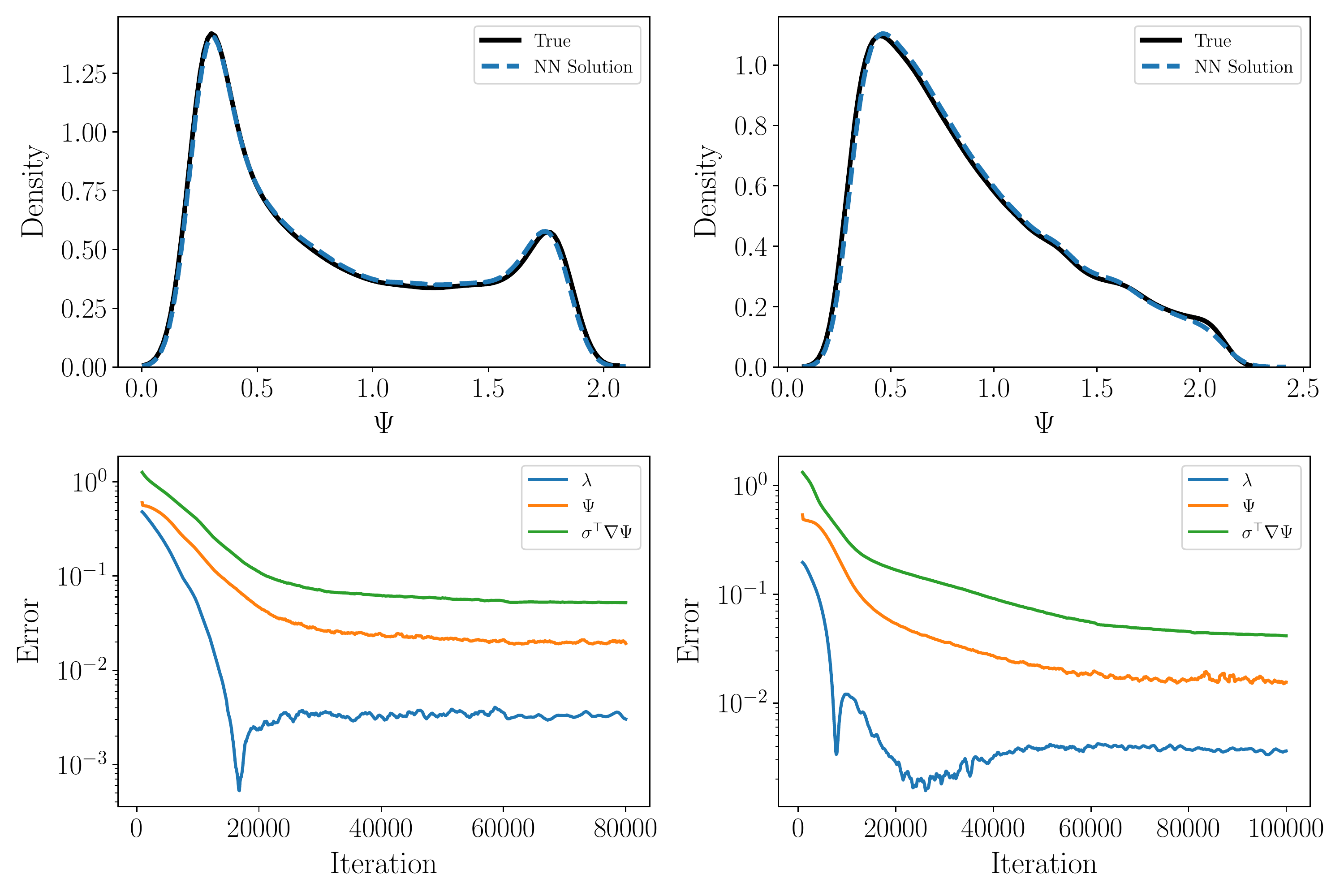}
    \caption{Top: density of $\Psi$ for Fokker-Planck equation with $d=5$ (left) and $d=10$ (right).
    Bottom: associated error curves in the training process with $d=5$ (left) and $d=10$ (right).}
    \label{fig:FP}
\end{figure*}

In this subsection we consider the linear Fokker-Planck operator 
$$\mathcal{L}\psi = -\Delta\psi - \nabla \cdot (\psi\nabla V) =  -\Delta\psi - \nabla V \cdot \nabla \psi - \Delta V \psi ,$$
where $V(x)$ is a potential function.
The smallest eigenvalue of $\mathcal{L}$ is $\lambda_1 = 0$ and the corresponding eigenfunction is $\Psi(x) = e^{-V(x)}$, which can be used to compute the error. We consider an example $V(x) = \sin\bigl(\sum_{i=1}^d c_i \cos(x_i)\bigr)$, where $x_i$ is the $i$-th coordinate of $x$, and $c_i$ takes values in $[0.1, 1]$. The function $V$ is periodic by construction. Figure~\ref{fig:FP} shows the density and error curves for the Fokker-Planck equation in $d=5$ and $d=10$. For $d=5$, the final errors of the eigenvalue, eigenfunction {(in $L^2$ and $L^{\infty}$)} and the scaled gradient are 3.08e-3, 2.91e-2, {1.25e-1} and 4.91e-2. For $d=10$, the final errors are 3.58e-3, 1.62e-2, {1.08e-1} and 4.10e-2.

\subsection{Linear Schr\"odinger equation} \label{LinearSdg}
In this subsection we consider the Schr\"odinger operator 
$$\mathcal{L}\psi = -\Delta\psi + V\psi,$$ 
where $V(x)$ is a potential function. Here we choose $V(x) = \sum_{i=1}^d c_i \cos(x_i)$, in which $c_i$ takes values in $[0, 0.2]$. With potential function being such a form, the problem is essentially decoupled. Therefore we are able to compute the eigenvalues and eigenfunctions in each dimension through the spectral method and obtain the final first eigenpair with high accuracy for comparison. The computation details are provided in Appendix \ref{spectrum}. Figure~\ref{fig:Sdg} shows the density and error curves for the Schr\"odinger equation in $d=5$ and $d=10$. For $d=5$, the final errors of the eigenvalue, eigenfunction {(in $L^2$ and $L^{\infty}$)} and the scaled gradient are 8.84e-4, 7.87e-3, {4.78e-2} and 2.30e-2. For $d=10$, the final errors are 1.40e-3, 1.19e-2, {7.48e-2} and 3.14e-2.

\begin{figure*}[t!]
    \centering
    \includegraphics[width=0.98\textwidth]{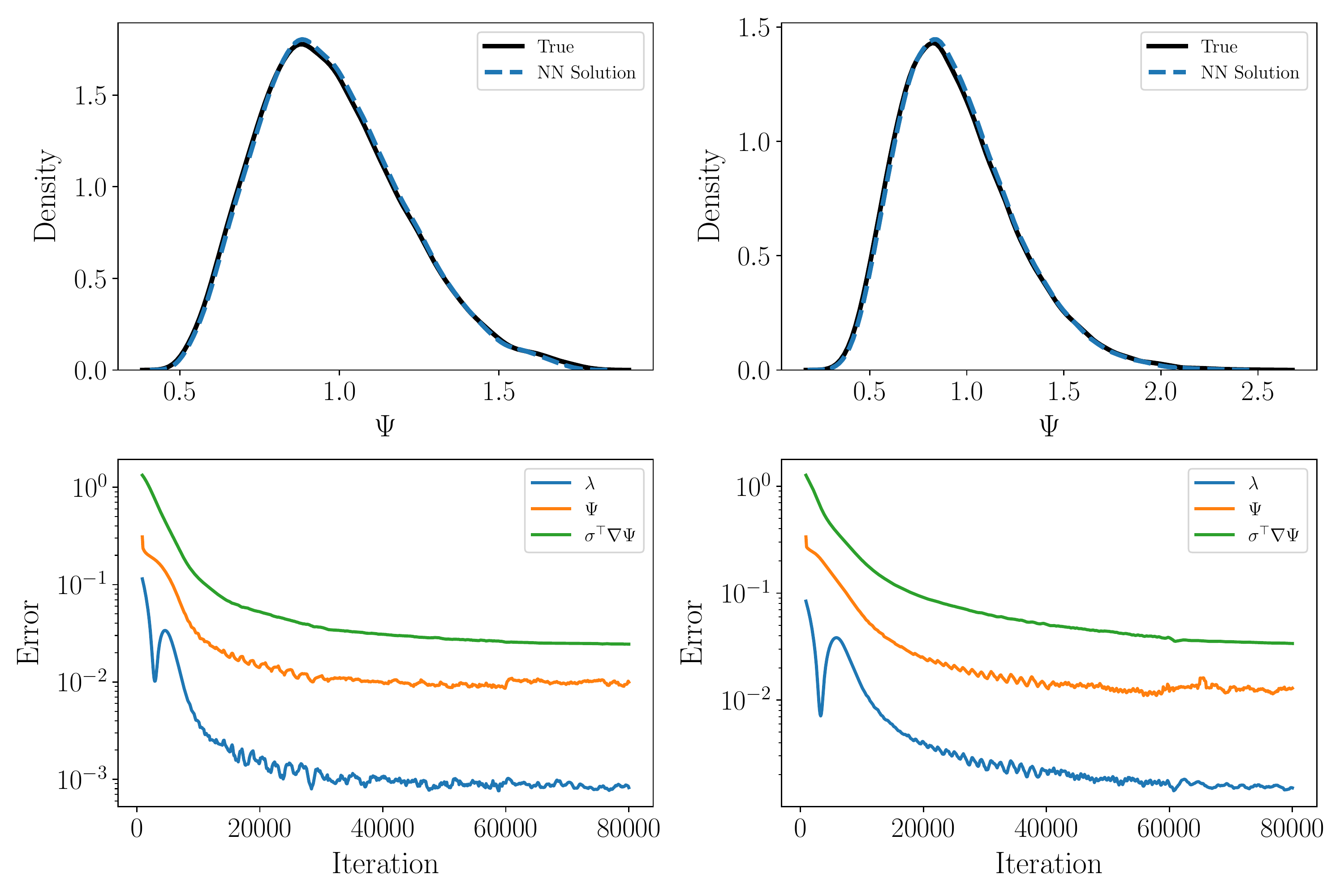}
    \caption{Top: density of $\Psi$ for linear Schr\"odinger equation with $d=5$ (left) and $d=10$ (right).
    Bottom: associated error curves in the training process with $d=5$ (left) and $d=10$ (right).}
    \label{fig:Sdg}
\end{figure*}

\subsection{Nonlinear Schr\"odinger equation}
\label{sec_cubic}
\begin{figure*}[t!]
    \centering
    \includegraphics[width=0.98\textwidth]{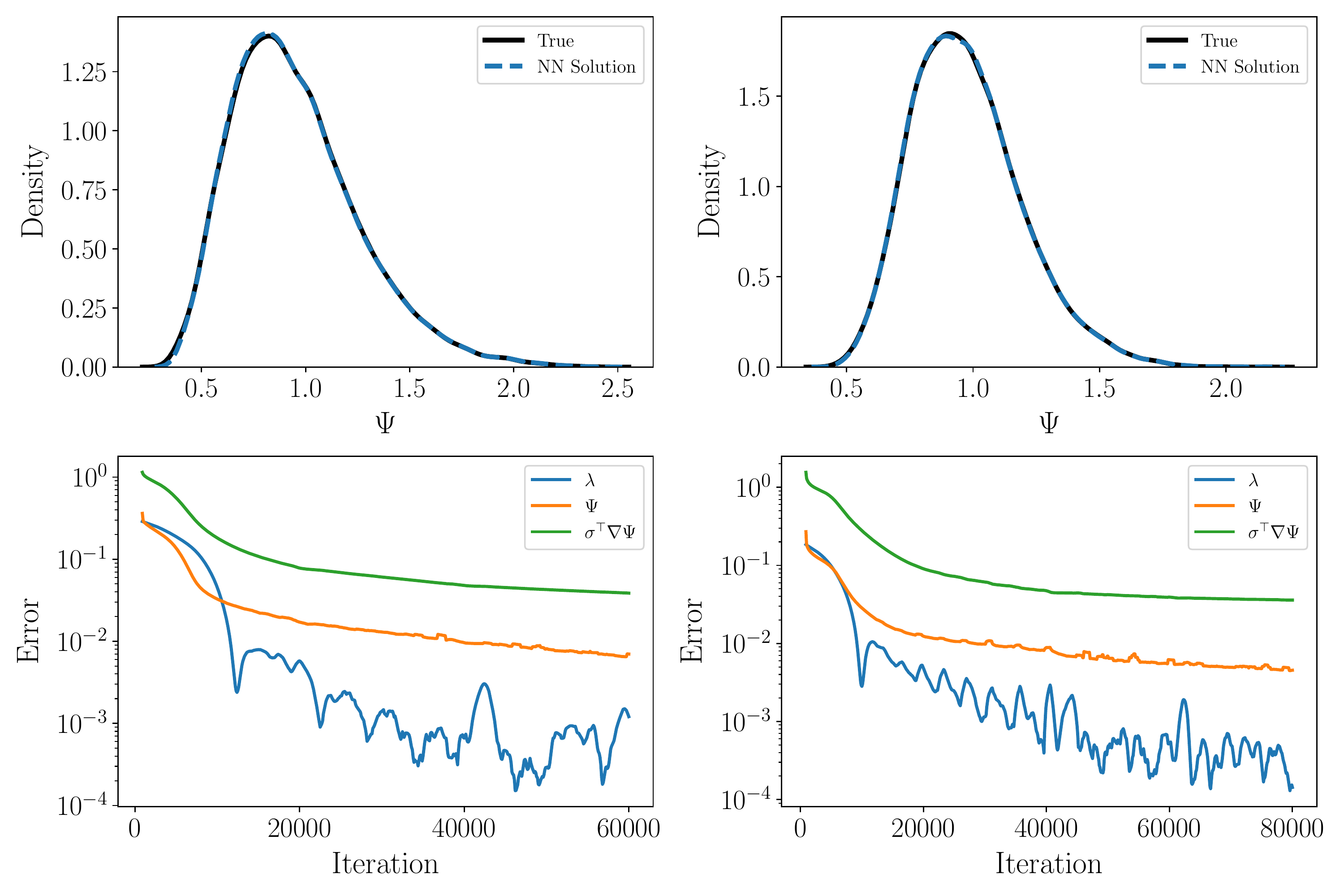}
    \caption{Top: density of $\Psi$ for nonlinear Schr\"odinger equation with $d=5$ (left) and $d=10$ (right). Bottom: associated error curves  in the training process  with $d=5$ (left) and $d=10$ (right).}
    \label{fig:NLSdg}
\end{figure*}

We finally consider a nonlinear Schr\"odinger operator with a cubic term, arising from the Gross-Pitaevskii equation~\cite{gross1961structure,pitaevskii1961vortex} for the single-particle wavefunction in a Bose-Einstein condensate: 
\begin{equation}
\label{eq:nlse}
\mathcal{L} \psi = - \Delta \psi + \epsilon \psi^3 + V \psi.
\end{equation}
Here we assume $\epsilon=1$ and consider a specific external potential
\begin{equation}
V(x) = -\frac{1}{c^2} \exp\left(\frac2d \sum_{i=1}^{d} \cos x_i\right) + \sum_{i=1}^{d} \left(\frac{\sin^2 x_i}{d^2} - \frac{\cos x_i}{d}\right) -3,
\end{equation}
such that $\lambda=-3, \Psi(x) = \exp(\frac{1}{d} \sum_{j=1}^{d} \cos(x_j))/c$ is an eigenpair of the operator \eqref{eq:nlse}. Here $c$ is a positive constant such that $\int_{\Omega} \Psi^2(x) \mathrm{d}x = |\Omega|$. 
{In this example we used $P=-5$ and $Q=5$ in the clip function \eqref{eqn:clip}}.
Figure~\ref{fig:NLSdg} shows the density and error curves for the nonlinear Schr\"odinger equation in $d=5$ and $d=10$. For $d=5$, the final errors of the eigenvalue, eigenfunction {in $L^2$ and $L^{\infty}$} and the scaled gradient are 1.53e-3, 8.07e-3, {6.58e-2} and 4.36e-2.
For $d=10$, the final errors are 2.6e-4, 4.60e-3, {3.83e-2} and 3.55e-2.

\subsection{An example for the second eigenpair}
\label{sec_doublewell}
{
In this subsection we reconsider the Schr\"odinger operator
$$\mathcal{L}\psi = -\Delta\psi + V\psi,$$ with the additional goal of finding the second eigenpairs. The potential function is double-well in each dimension: $V(x) = \sum_{i=1}^d A_i \cos(2x_i)$.
The reference solutions are solved by the same numerical method as in Section \ref{LinearSdg}.
}

\begin{figure*}[t!]
    \centering
    \includegraphics[width=0.98\textwidth]{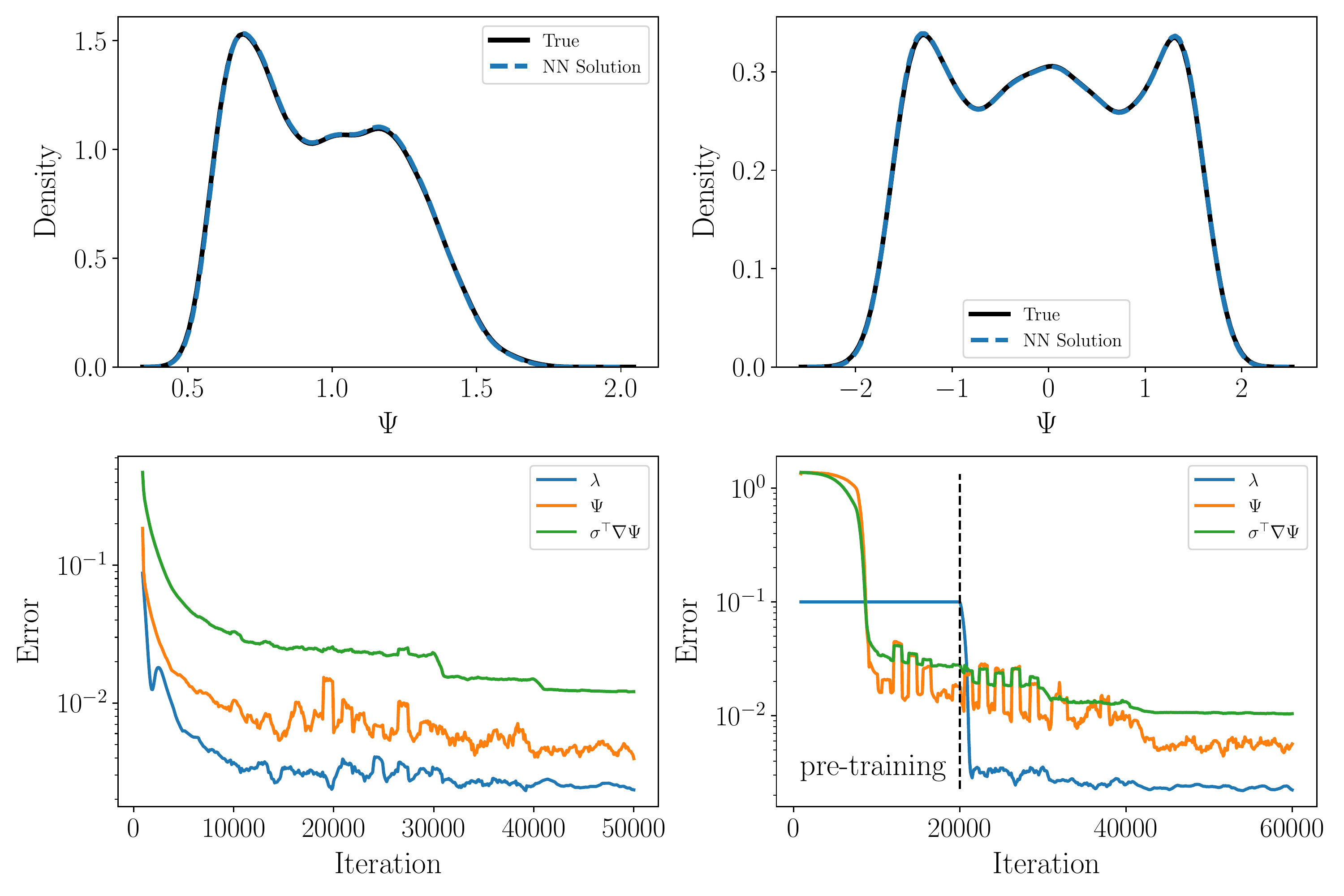}
    \caption{Top: density of the first (left) and second (right) eigenfunction $\Psi$ for linear Schr\"odinger equation $(d=10)$ with double-well potential. Bottom: associated error curves in the training process for the first (left) and second (right) eigenfunction. In the error curve on the right, the error for eigenvalue keeps unchanged in the first 20000 steps as we fixed it in this pre-training procedure.}
    \label{fig:DWSdg2}
\end{figure*}

{We first consider the cases when the eigenvalues are well-separated. Assuming $d=10$, $A_1=1.5$ and $A_i=0.2$ for $2\leq i \leq 10$, the associated eigen-gaps are $\lambda_2 - \lambda_1 =$ 4.52e-1 and $\lambda_3 - \lambda_2$ = 4.52e-1. If we follow the training procedure described previously, we are able to find the first eigenpair with errors of 2.34e-3, 3.95e-3, 2.37e-2 and 1.21e-2. The left column in Figure~\ref{fig:DWSdg2} shows the density and error curves.}

{On the other hand, our method is also able to find the second eigenpair given some mild prior estimate of the eigenvalue. Suppose that we have an approximation $\overline{\lambda}$ of the true second eigenvalue $\lambda_2$. We can firstly fix this approximated eigenvalue and train the neural networks only with some steps. With such a pre-training procedure, we expect that the neural networks would reach a reasonable approximation of the second eigenfunction and its scaled gradient. Then, we train both the eigenvalue and the neural networks for the eigenfunction until convergence.
The right column in Figure~\ref{fig:DWSdg2} shows the density and error curves of the second eigenpair, obtained by following the described procedure. In this example, $\overline{\lambda} = \lambda_2 + 0.1$, and the final errors are 2.23e-3, 5.65e-3, 5.08e-2 and 1.04e-2. 
}

\begin{figure*}[t!]
    \centering
    \includegraphics[width=0.98\textwidth]{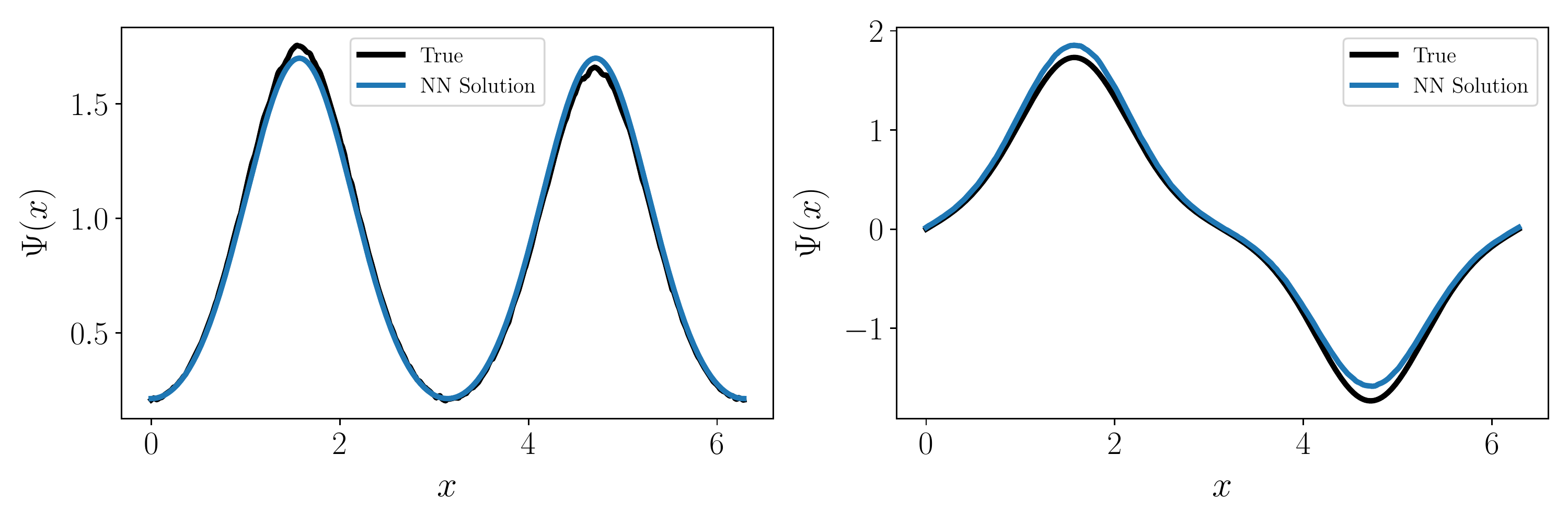}\\
    \caption{Plot of the eigenfunctions for one-dimensional degenerate linear Schr\"odinger problem with double-well potential. Left: first eigenfunction. Right: second eigenfunction.}
    \label{fig:DW2Sdg}
\end{figure*}

{
We remark that the eigenvalue problem becomes more challenging when it is nearly degenerate, i.e., the first and second eigenvalues are close to each other. 
This is a common phenomenon for various numerical methods, and our algorithm is no exception.
Suppose we consider a one-dimensional problem with potential $V(x)=5\cos(2x)$. The first and second eigenvalues are $-2.153$ and $-2.076$ respectively, with a small gap 8.7e-2.
If we train the model for the first eigenpair like before,
the obtained eigenfunction is plotted on
the left of Figure \ref{fig:DW2Sdg}.
If we solve the second eigenpair with the additional pre-training procedure introduced above, we still cannot get the second eigenfunction approximately, even with a fixed true second eigenvalue.
The results can be improved if we furthermore have certain approximation to the second eigenfunction. For instance, if we initialize the neural networks for the eigenfunction as $\psi_{init} = \psi_2 + \epsilon \psi_1$ where $\psi_1$ and $\psi_2$ are the first and second eigenfunctions respectively and $\epsilon$ denotes the degree of perturbation. 
We take $\epsilon = 0.3$ in our numerical experiment, and we remark that the direction of perturbation is not necessarily in $\psi_1$. 
Such initialization can be achieved through a supervised learning procedure.
With this initialization, the final solution to the second eigenfunction is plotted on the right of Figure~\ref{fig:DW2Sdg}.
}

{
According to the numerical results of both the well-separated and degenerate problems, we find that better initialization of eigenvalue or eigenfunctions can usually improve results. 
For the degenerate problem, even in the one-dimensional case,
both the first and second eigenpairs are difficult to solve accurately since the final solution may involve a mix of two eigenfunctions.
}

\section{Conclusion and future works}
\label{sec:conclusion}
In this paper, we propose a new method to solve eigenvalue problems in high dimensions using neural networks. Our method is able to compute both eigenvalues and corresponding eigenfunctions (with gradients) with high accuracy. 

There are several natural directions for future work. First, to apply our methodology to quantum many-body systems, we need to respect the permutation symmetry in our ansatz for the wavefunctions. Previous works ~\cite{Han2019solving,Choo2019fermionic,Hermann2019deep, Pfau2019abinitio} have proposed various flexible neural-network ansatz to incorporate in the symmetry, which can be combined with our approach. Moreover, in DMC, importance sampling techniques are often essential to improve the accuracy. In our context, this means to choose a better underlying diffusion process guided by a trial wavefunction depending on the problem. Last, the scalability of the method has to be tested on larger systems beyond the toy numerical examples in this work. 

On the theoretical aspects, the understanding of the stability and convergence of the proposed method is a fascinating future direction. While the general analysis might be quite difficult given the highly nonlinear approximation induced by the neural networks and also the complicated optimization strategy, some perturbative analysis, especially in the linearized regime, might be possible. We will leave these to future works. 
 
\medskip
\noindent\textbf{Acknowledgement.} The work of JL and MZ is supported in part by National Science Foundation via grant DMS-1454939 and DMS-2012286. The authors are grateful for computing time at the Terascale Infrastructure for Groundbreaking Research in Science and Engineering (TIGRESS) of Princeton University.

\bibliography{ref}{}
\bibliographystyle{unsrt}

\begin{appendices}
\section{Spectrum method for linear Schr\"odinger equation}
\label{spectrum}
Suppose that the potential function in the linear Schr\"odinger operator $\mathcal{L} = -\Delta + V$ is decoupled with the form $V(x) = \sum_{j=1}^d c_j \cos(x_j)$, then we can solve the corresponding eigenvalue problem in a decoupled way. Specifically, assume we can solve the one-dimensional eigenvalue problem
\begin{equation} \label{eqn:1dSdg}
-\Psi_j''(x) + c_j \cos(x) \Psi_j(x) = \lambda_j \Psi_j(x), ~~~ x \in [0,2\pi].
\end{equation}
Then one can easily verify that $\lambda = \sum_{j=1}^{d} \lambda_j$ and 
\begin{equation}
\Psi(x) = \prod_{j=1}^{d} \Psi_j(x_j)
\end{equation}
together define an eigenpair of the original high-dimensional Schr\"odinger operator.

To solve \eqref{eqn:1dSdg}, we can employ the classical spectrum method. For a fixed $N \in \mathbb{N}$, assume that
\begin{equation}
\label{eqn:fourier}
\Psi_j(x) = \sum_{m=-N}^{N} a^j_m e^{imx},
\end{equation}
then
\begin{equation}
\Psi'_j(x) = \sum_{m=-N}^{N} m a^j_m e^{imx}i.
\end{equation}
Let $\varphi_n(x) = e^{inx}$ ($n = -N, \cdots, N$) be the test functions. By \eqref{eqn:1dSdg} and periodicity, we have
\begin{equation} \label{eqn:integral}
\int_{0}^{2\pi} (\Psi'_j(x) \varphi_n'(x) + c_j \cos(x) \Psi_j(x) \varphi_n(x)) \mathrm{d}x = \lambda_j \int_{0}^{2\pi} \Psi_j(x) \varphi_n(x) \mathrm{d}x.
\end{equation}
Since $\int_{0}^{2\pi} e^{imx} e^{inx} \mathrm{d}x = 2\pi \delta_{m+n}$ and  $\int_{0}^{2\pi} \cos(x) e^{imx} e^{inx} \mathrm{d}x = \frac12 \int_{0}^{2\pi} (e^{i(m+n+1)x} + e^{i(m+n-1)x}) \mathrm{d}x = \pi (\delta_{m+n+1} + \delta_{m+n-1})$, the left- and right-hand sides of \eqref{eqn:integral} become
\begin{equation}
\begin{aligned}
&~\int_{0}^{2\pi} (\Psi'_j(x) \varphi_n'(x) + c_j \cos(x) \Psi_j(x) \varphi_n(x)) \mathrm{d}x \\
=&~ \sum_{m=-N}^{N} a_m^j \int_{0}^{2\pi} (-mn e^{imx} e^{inx} + c_j \cos(x) e^{imx} e^{inx}) \mathrm{d}x \\
=&~ \sum_{m=-N}^{N} a_m^j \pi (-2mn \delta_{m+n} + c_j \delta_{m+n+1} + c_j \delta_{m+n-1})\\
=&~ \pi (2n^2 a_{-n} + c_j a_{-n-1} + c_j a_{-n+1}),
\end{aligned}
\end{equation}
(assuming $a_m=0$ for $|m|>N$) and
\begin{equation}
\int_{0}^{2\pi} \Psi_j(x) \varphi_n(x) \mathrm{d}x = 2 \pi a_{-n},
\end{equation}
respectively. Therefore, equation \eqref{eqn:integral} can be rewritten in the matrix form:
\begin{align}
\pi \begin{bmatrix}
2N^2&c_j&&&&&\\ c_j&\ddots&\ddots&&&&\\ &\ddots&2&\ddots&&&\\ &&\ddots&0&\ddots&&\\ &&&\ddots&2&\ddots&\\ &&&&\ddots& \ddots&c_j\\ &&&&&c_j&2N^2
\end{bmatrix} 
\begin{bmatrix}
a^j_{-N} \\ \vdots \\ \vdots \\ \vdots \\ a^j_{N}
\end{bmatrix} = 
2 \pi \lambda_j  
\begin{bmatrix}
a^j_{-N} \\ \vdots \\ \vdots \\ \vdots \\ a^j_{N}
\end{bmatrix}.
\label{eqn:spectrum_matrix}
\end{align}
This is a standard eigenvalue problem in numerical algebra. Suppose $\tilde{\lambda}_j$ and $(a_{-N}, \dots, a_N)$ is the eigenvalue and associated eigenvector of the matrix in \eqref{eqn:spectrum_matrix}, then $\lambda_j = \tilde{\lambda}_j/2$ is the approximated eigenvalue in \eqref{eqn:1dSdg}, and \eqref{eqn:fourier} provides the associated eigenfunction, which is equivalent to a real eigenfunction up to a complex constant.

\section{Hyperparameters in the numerical example}
\label{hyperparameter}
We first report hyperpameters commonly used in all three numerical examples and then list in Tables \ref{tab:FP}-\ref{tab:cubicSDG} those specific to the examples.
In all three examples, the order of the trigonometric basis $M=5$, the constant $Z_0=2$ for regularizing the normalization factor, the weight parameters in the loss $\eta_1=1000, \eta_2=20, \eta_3=100$.
In the following tables, the structures of the neural networks are represented by vectors, whose elements denote the number of nodes within each layer. The learning rate and moving average decay $\gamma_{\ell}$ are both piecewise constant, whose values and boundaries are given separately. For example, in $5$-dimensional Fokker-Planck problem, the learning rate is \num{1e-4} for the first $30000$ steps, \num{5e-5} from the $30001$-st to the $60000$-th step and 1e-5 after the $60000$-th step. The moving average decay $\gamma_{\ell}$ is defined similarly, with the same boundaries. 

{We remark that the choice of the terminal time $T$ is a trade-off between discretization errors and training errors. For a fixed number of time steps $N$, a large $T$ will result in large discretization errors. On the other hand, if a small $T$ is used, the discrepancy between $\mathcal{P}_T^\lambda\Psi$ and $\Psi$ is less significant when a wrong $\Psi$ is used (think about the extreme case that $T = 0$, any $\Psi$ would give $0$ loss), which causes difficulty in the optimization of the parameters.}

{We also remark that the choice of the size of the neural networks is a trade-off between the approximation accuracy and computational cost. Three is chosen as a moderate depth while the widths are chosen to guarantee enough approximation capability. We choose ReLU as the activate function to save the computation cost of backpropagation in the calculation of derivatives, without a sacrifice of accuracy. The learning rates are chosen to be non-increasing piecewise constant according to common practice.}
\newline

\begin{table}[htbp]
\centering
\begin{tabular}{c | c c} 
\toprule
Parameters & $d=5$ & $d=10$ \\ 
\midrule
terminal time $T$ & $0.2$ & $0.2$ \\ 
number of time intervals $N$ & $80$ & $120$ \\
structure of neural networks & [$80$, $80$, $80$] & [$300$, $300$, $300$] \\
number of iterations & $80000$ & $100000$\\
learning rate & [\num{1e-4}, \num{5e-5}, \num{1e-5}] & [\num{5e-5}, \num{2e-5}, \num{1e-5}]\\
moving average decay $\gamma_{\ell}$ & [$0.2$, $0.5$, $0.9$] & [$0.2$, $0.5$, $0.9$]\\
piecewise constant boundaries & [$30000$,$60000$] & [$60000$, $80000$]\\
batch size $K$ & $1024$ & $1024$\\
\bottomrule
\end{tabular}
\caption{Parameters for Fokker-Planck eigenvalue problems in Section \ref{secFP}.}
\label{tab:FP}
\end{table}

\begin{table}[htbp]
\centering
\begin{tabular}{c | c c} 
\toprule
Parameters & $d=5$ & $d=10$ \\ 
\midrule
terminal time $T$ & $0.3$ & $0.3$ \\ 
number of time intervals $N$ & $80$ & $120$ \\
structure of neural networks & [$80$, $80$, $80$] & [$300$, $300$, $300$] \\
number of iterations & $80000$ & $80000$\\
learning rate & [\num{1e-4}, \num{5e-5}, \num{1e-5}] & [\num{5e-5}, \num{5e-5}, \num{1e-5}]\\
moving average decay $\gamma_{\ell}$ & [$0.2$, $0.5$, $0.9$] & [$0.2$, $0.5$, $0.9$]\\
piecewise constant boundaries & [$30000$, $60000$] & [$40000$, $60000$]\\
batch size $K$ & $1024$ & $1024$\\
\bottomrule
\end{tabular}
\caption{Parameters for linear Schr\"odinger eigenvalue problems in Section \ref{LinearSdg}.}
\label{tab:linearSDG}
\end{table}

\begin{table}[htbp]
\centering
\begin{tabular}{c | c c} 
\toprule
Parameters & $d=5$ & $d=10$ \\ 
\midrule
terminal time $T$ & $0.2$ & $0.3$ \\ 
number of time intervals $N$ & $120$ & $200$ \\
structure of neural networks & [$80$, $80$, $80$] & [$300$, $300$, $300$] \\
number of iterations & $60000$ & $80000$\\
learning rate & [\num{5e-5}, \num{2e-5}, \num{1e-5}] & [\num{5e-5}, \num{2e-5}, \num{1e-5}]\\
moving average decay $\gamma_{\ell}$ & [$0.2$, $0.9$, $0.99$] & [$0.1$, $0.9$, $0.99$]\\
piecewise constant boundaries & [$20000$, $40000$] & [$40000$, $60000$]\\
batch size $K$ & $2048$ & $2048$\\
\bottomrule
\end{tabular}
\caption{Parameters for nonlinear Schr\"odinger eigenvalue problems in Section \ref{sec_cubic}}
\label{tab:cubicSDG}
\end{table}

\begin{table}[htbp]
\centering
\begin{tabular}{c | c c} 
\toprule
state & first eigenpair & second eigenpair \\ 
\midrule
terminal time $T$ & $0.2$ & $0.2$ \\ 
number of time intervals $N$ & $320$ & $320$ \\
structure of neural networks & [$200$, $200$, $200$] & [$200$, $200$, $200$] \\
number of iterations & $50000$ & $50000$\\
learning rate & [\num{5e-4}, \num{1e-4}, \num{1e-5}] & [\num{5e-4}, \num{1e-4}, \num{1e-5}]\\
moving average decay $\gamma_{\ell}$ & [$0.1$, $0.2$, $0.9$] & [$0.1$, $0.9$, $0.99$]\\
piecewise constant boundaries & [$30000$, $40000$] & [$30000$, $40000$]\\
batch size $K$ & $2048$ & $2048$\\
\bottomrule
\end{tabular}
\caption{Parameters for well-separated linear Schr\"odinger eigenvalue problems with $d=10$ in Section~\ref{sec_doublewell}.}
\label{tab:DWSDG}
\end{table}

\begin{table}[htbp]
\centering
\begin{tabular}{c |c c} 
\toprule
state & first eigenpair & second eigenpair \\ 
\midrule
terminal time $T$ & $0.2$ & $0.2$ \\ 
number of time intervals $N$ & $80$ & $80$ \\
structure of neural networks & [$40$, $40$] & [$40$, $40$] \\
number of iterations & $6000$ & $6000$\\
learning rate & [\num{5e-4}, \num{1e-4}, \num{1e-5}] & [\num{5e-4}, \num{1e-4}, \num{1e-5}] \\
moving average decay $\gamma_{\ell}$ & [$0.1$, $0.2$, $0.9$] & [$0.1$, $0.2$, $0.9$] \\
piecewise constant boundaries & [$2000$, $4000$] & [$2000$, $4000$]\\
batch size $K$ & $512$ & $512$\\
\bottomrule
\end{tabular}
\caption{Parameters for degenerate linear Schr\"odinger eigenvalue problems with $d=1$ in Section~\ref{sec_doublewell}.}
\label{tab:double-welld1}
\end{table}
\end{appendices}
\end{document}